\pdfoutput=1

\documentclass[11pt]{article}

\usepackage{acl}

\usepackage{times}
\usepackage{latexsym}

\usepackage[T1]{fontenc}

\usepackage[utf8]{inputenc}

\usepackage{microtype}

\usepackage{inconsolata}

\usepackage{amsmath}
\usepackage{mathtools}
\usepackage{amsfonts}
\usepackage{xcolor}
\usepackage{multirow}

\title{Scaling Up Summarization: Leveraging Large Language Models for Long Text Extractive Summarization}

\author{Mehdi Debiane \\
  Sanofi \\
  \texttt{mehdi.debiane@sanofi.com} \\ \And
  Léo Hemamou \\
  Sanofi \\
\texttt{leo.hemamou@sanofi.com} \\}
\begin{document}
\maketitle

\begin{abstract}
In an era where digital text is proliferating at an unprecedented rate, efficient summarization tools are becoming indispensable. While Large Language Models (LLMs) have been successfully applied in various NLP tasks, their role in extractive text summarization remains underexplored. This paper introduces EYEGLAXS (Easy Yet Efficient larGe LAnguage model for eXtractive Summarization), a framework that leverages LLMs, specifically LLAMA2-7B and ChatGLM2-6B, for extractive summarization of lengthy text documents. Instead of abstractive methods, which often suffer from issues like factual inaccuracies and hallucinations, EYEGLAXS focuses on extractive summarization to ensure factual and grammatical integrity. Utilizing state-of-the-art techniques such as Flash Attention and Parameter-Efficient Fine-Tuning (PEFT), EYEGLAXS addresses the computational and resource challenges typically associated with LLMs. The system sets new performance benchmarks on well-known datasets like PubMed and ArXiv. Furthermore, we extend our research through additional analyses that explore the adaptability of LLMs in handling different sequence lengths and their efficiency in training on smaller datasets. These contributions not only set a new standard in the field but also open up promising avenues for future research in extractive text summarization.
\end{abstract}

\section{Introduction}

In the era of information overload, text summarization has emerged as a critical tool for distilling essential information from expansive text documents. This paper focuses on automatic text summarization, which can be broadly categorized into two paradigms: abstractive and extractive methods. Abstractive methods, despite their ability to generate flexible and creative summaries, often grapple with issues of grammatical inaccuracy and factual inconsistencies, commonly referred to as "hallucinations" \cite{bishop2022gencomparesum,Ji_2023hallucination,zhang2023sirens}. These challenges are exacerbated when summarizing long texts and can be particularly detrimental in critical applications such as healthcare, scientific research, and legislation. In contrast, extractive summarization offers a more reliable approach by selecting pertinent sentences directly from the source text, thereby ensuring grammatical and factual integrity. Traditionally, this task has been framed as a sentence classification problem and has predominantly employed encoder-only pre-trained models \cite{liu-lapata-2019-text,cho-etal-2022-toward,bian2023gosum}.
Despite the promising capabilities of Large Language Models (LLMs) in various NLP tasks, their potential in extractive summarization remains largely untapped. This oversight is partly due to the computational challenges and fine-tuning limitations associated with these sizable models. However, recent advancements in long sequence processing for decoder-only models offer a glimmer of hope for harnessing LLMs in this context.
To bridge this gap, our paper introduces EYEGLAXS (Easy Yet Efficient larGe LAnguage model for eXtractive Summarization), a system that leverages the power of LLMs—specifically LLAMA2-7B\cite{touvron2023llama} and ChatGLM2-6B \cite{zeng2022glm}. We employ Flash Attention 2 \cite{dao2023flashattention2} and Parameter-Efficient Fine-Tuning (PEFT)\cite{lialin2023scaling} techniques to mitigate some of the challenges associated with using LLMs. Our contributions are manifold: we not only propose a novel method for employing LLMs in long extractive summarization tasks but also demonstrate their competitive performance against state-of-the-art methods. We further explore the adaptability of LLMs in handling varying sequence lengths and investigate their training efficiency on smaller datasets. Lastly, we delve into the issue of position bias inherent in LLMs.






\section{Litterature Review}

\subsection{Long Extractive Text Summarization}

In the literature, the task of extractive text summarization is predominantly approached as a sentence classification problem. In this framework, models are trained to predict a label for each sentence in the input document to determine whether or not the sentence should be included in the generated summary. Most state-of-the-art methods leverage pre-trained transformer models that are adapted for natural language understanding tasks. One of the pioneering works in this area slightly modified BERT's architecture by incorporating a priori information on sentence splitting and adding layers of inter-sentence transformers before feeding them into the classifier for prediction \cite{liu-lapata-2019-text}. To address the issue of limited context size, various transformer architectures have been proposed to mitigate the quadratic complexity problem associated with self-attention computation. For instance, Longformer \cite{beltagy2020longformer} and Bigbird \cite{Zaheer2020bigbird} employ attention sparse methods such as sliding windows to handle longer sequences. Building on these architectures, many systems introduce additional mechanisms that exploit the unique characteristics of documents to improve performance on long sequences. For example, the work by \cite{xiao-carenini-2019-extractive} generates different representations that consider both local and global contexts. The system described in \cite{ruan-etal-2022-histruct} explicitly incorporates hierarchical information by using section titles and the hierarchical position of sentences to enrich representations. Similarly, \cite{bian2023gosum} represents the hierarchical structure of the text through a heterogeneous graph of sentences and sections, while integrating reinforcement learning with a graph neural network. The approach by \cite{cho-etal-2022-toward} aims to discover the latent structure of the document by jointly optimizing a secondary task of section segmentation alongside sentence extraction. Moreover, \cite{xie2022pre} incorporates domain-specific knowledge into the model by using adapters to infuse medical expertise. The abstractive-extractive approach has also been explored. For instance, \cite{bishop2022gencomparesum} generates an abstractive summary that later guides the extraction of salient sentences, irrespective of document length. Most of these approaches employ pre-trained RoBERTa \cite{DBLProberta} as the backbone model. However, this comes with limitations, such as the complexity of learning new positional encoding tokens and the relatively small number of parameters and tokens encountered during the pre-training stage, especially when compared to larger language models.

\subsection{Large Language Models}

Over the past few years, pre-trained Large Language Models (LLMs) have transitioned from being virtually unknown to becoming pervasive in the realm of machine learning. Their widespread adoption is largely due to their proven effectiveness in addressing zero-shot and few-shot learning challenges. They have been successfully applied to tasks such as abstractive summarization and translation of short documents. While these models excel in generative tasks, their application to extractive tasks poses greater challenges. A common workaround is to transform an extractive task into a text generation task by utilizing cloze prompting templates \cite{pretrainprompt}. While this technique is well-suited for simpler tasks like sentiment detection, its complexity escalates for tasks with more intricate scoring systems, such as named entity recognition, and becomes nearly unfeasible for tasks like extractive summarization \cite{2023arXiv230410428WGPTNER}. Additionally, these models are prone to hallucination issues, limiting their applicability in critical fields like healthcare \cite{kaddour2023challenges}. Although zero-shot approaches to extractive summarization have been explored \cite{zhang2023extractivegpt}, to the best of our knowledge, no attempts have been made to evaluate the fine-tuning of these models specifically for extractive summarization. Furthermore, it is plausible that the representations learned by these LLMs are richer than those learned by encoder models, owing to their larger number of parameters and more extensive pretraining data \cite{ni-etal-2022-sentence}. Finally, efforts have been made to significantly increase the context size of these LLMs, notably through the use of scalable positional encodings \cite{Rotaryrope}, extending the initial pretraining context length \cite{chen2023extendingPI}, or optimizing attention computation at the GPU memory level \cite{dao2022flashattention,dao2023flashattention2}.

\subsection{Parameter-Efficient Fine-Tuning methods}


Large Language Models (LLMs) are increasingly being utilized to achieve state-of-the-art performance in various NLP tasks, capturing the attention of both researchers and industry professionals \cite{bubeck2023sparks}. However, these models come with significant computational and memory requirements for training from scratch. Recent iterations of these models often boast more than 70 billion parameters \cite{touvron2023llama,zeng2022glm}, making the fine-tuning process highly resource-intensive.
To address this challenge, a new family of techniques known as Parameter-Efficient Fine-Tuning (PEFT)\cite{lialin2023scaling} has been introduced. These methods advocate for the training of a relatively small number of additional parameters, which, in comparison to the overall size of the models, represent only a fraction. This approach substantially reduces both storage and computational costs. Among the PEFT techniques, prompt tuning, prefix tuning \cite{liu-etal-2022-ptuning}, and LoRA \cite{hu2022LoRA} have garnered significant attention. Models trained using PEFT methods have demonstrated performance levels comparable to those achieved through full fine-tuning \cite{lialin2023scaling}. 
In this article, we propose to fine-tune pre-trained LLMs using LoRA for the specific task of extractive summarization.

\begin{figure}
    \centering
    \includegraphics[width=1\linewidth]{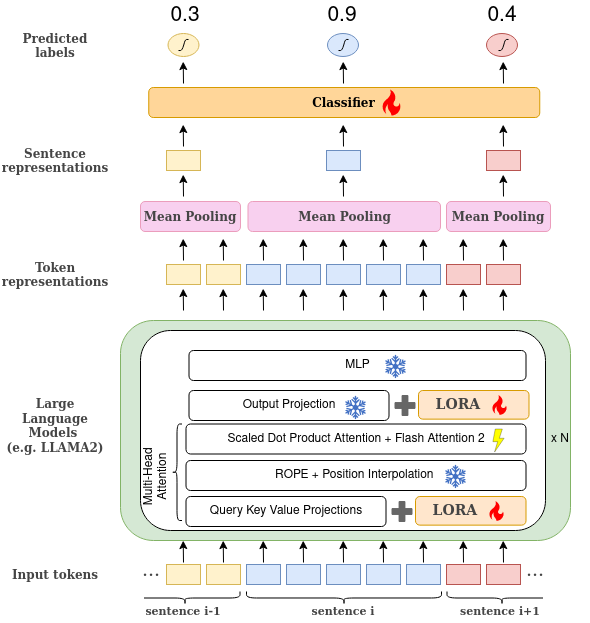}
    \caption{The overall framework of EYEGLAXS. Residual connections and normalizations do not appear for better readability. Snowflake logo means that weights are frozen, while Fire logo means that weights are trainable.}
    \label{fig:fig1}
\end{figure}

\section{Method}

\subsection{Task Definition}

We define extractive summarization as a sentence classification problem. Let note $D = \{s_1,s_2, \dots,s_n\}$ the document D consisting of $n$ sentences. The extractive summarization task aims to predict labels $\hat{y}_i \in (0,1)$ for each \( i \in [0, n] \) where $\hat{y}_i=1$ and $\hat{y}_i=0$ means the sentence should be included , or not, in the summary respectively. As datasets only contain document-abstract pairs, in order to train these models in a supervised manner, reference extractive abstracts are generated using methods such as greedy algorithms with the objective to maximize a chosen metric, mostly ROUGE scores \cite{kedzie-etal-2018-content,lin-2004-rouge}. This allows associating a label to each sentence of the original document based on their inclusion in these oracle summaries.
These oracle summaries represent an upper bound for the expected performances of extractive models. Thus a label $y_i\in (0,1)$, is associated to each sentence of the text depending on its inclusion in the oracle.

We propose a system called EYEGLAXS (Easy Yet Efficient larGe LAnguage model for eXtractive Summarization), a system based on Large Language Models for the long text extractive summarization task described in Figure \ref{fig:fig1}. 

\subsection{Choice of Large Language Models}

Large Language Models (LLMs) are notably resource-intensive to train from scratch, making it common practice to leverage pre-existing architectures for new applications. To improve both reproducibility and evaluation, we employ fine-tuning techniques on models that have publicly available checkpoints. A crucial aspect of our selection process is the choice of models that feature extendable positional encoding mechanisms \cite{Rotaryrope, press2021trainalibi}. We also ensure that the selected positional encoding can be efficiently parallelized to benefit from Flash Attention 2, which offers efficient GPU memory consumption \cite{dao2022flashattention, dao2023flashattention2}. As a result, models relying on relative positional encoding—such as those based on the T5 architecture—are not suitable for our use case. Likewise, models with learned positional encoding, like XLM \cite{goyal-etal-2021-largerxlm}, pose scalability challenges unless new positional encodings are retrained.
Guided by these considerations, we have narrowed our evaluation to two distinct LLMs: first, LLAMA2 \cite{touvron2023llama}, a decoder-only model that has gained widespread adoption; and second, ChatGLM2 \cite{zeng2022glm}, a prefix decoder model that shows promise for superior information integration through the use of bidirectional attention mechanisms.
More specifically, we assess the performance of the pre-trained long-sequence instruct-based models LLAMA2-7B-32K-Instruct\footnote{https://huggingface.co/togethercomputer/Llama-2-7B-32K-Instruct} and ChatGLM2-6B-32K\footnote{https://huggingface.co/THUDM/chatglm2-6b-32k}, both available on HuggingFace. The former has been fine-tuned for long-context summarization using the BookSum dataset, while the latter excels in conversational contexts. During our experiments, the focus was primarily on evaluating the feasibility of fine-tuning these models using LoRA. We did not place particular emphasis on the instructions used during training. Instead, we adhered to minimalist prompts, which were formed by concatenating the sentences of the input documents. This approach respected the format used by each model and did not include any additional instructions.

\subsection{Transformer}

Large Language Models (LLMs) are commonly built using transformer architectures. Architectures such as LLAMA2 \cite{touvron2023llama} and GLM \cite{zeng2022glm} consist of an initial embedding input layer followed by multiple decoder layers. Each of these decoder layers comprises Query-Key-Value (QKV) Projection Layers, Rotary Position Encoding, a self-attention module, an output projection, a multi-layer perceptron, residual connections, and normalization. For the sake of readability, we focus solely on our modifications to the original architecture in the areas of QKV Projection Layers, Rotary Position Encoding, self-attention module, and output projection.


\subsection{Query Key Value Projection Layers with LoRA} 
\label{sec:qkvLoRA}
This module transforms the input token $x_m$ with position $m$ into a trio of queries, keys, and values—represented as $\{q_m, k_m, v_m\}$ —through linear projection layers with corresponding weight matrices $\{Wq , Wk , Wv\}$. These matrices are later used to compute attention values. We apply LoRA on these specific matrices. More precisely, for a pre-trained matrix $W \in \mathbb{R}^{d\cdot p}$ , we add low rank adapter $\delta W$ such as :

\begin{equation}
\begin{split}
q_m = (W_q+\delta W_q)x_m = (W + B_q A_q)x_m \\
k_m = (W_k+\delta W_k)x_m = (W + B_k A_k)x_m \\
v_m = (W_v+\delta W_v)x_m = (W + B_v A_v)x_m \\
\end{split}
\end{equation}
where $B_{\{q,k,v\}} \in \mathbb{R}^{d\times r}$ , $A_{\{q,k,v\}} \in \mathbb{R}^{r\times p}$, $d$ is the dimension of $x_m$ and with $r \ll min(d,p)$ . The idea is that the weight updates in pre-trained models have a low intrinsic rank during adaptation. Thus, during training, $W_{\{q,k,v\}}$ weights are frozen and the number of trainable parameters (i.e. $B_{\{q,k,v\}}$ and $A_{\{q,k,v\}}$ ) are drastically reduced compared to full fine-tuning setting.

\subsection{Rotary Positional Encoding }

We employ architectures using a novel positional encoding scheme called Rotary Positional Embeddings (RoPE)\cite{Rotaryrope}. RoPE is a distinctive form of positional embedding used in Transformer models to encode the absolute and relative positional information of tokens within a sequence. Moreover, RoPE is valued for its flexibility to expand to any sequence lengths \cite{Rotaryrope}. 

The mathematical intuition behind RoPE aims to devise a positional encoding function $f(x,m)$ for a token $x$ at position $m$ such that for a query vector $q_m$ and a key vector $k_l$ at positions $m$ and $l$ respectively, the inner product between $f(q_m,m)$ and $f(k_l,l)$ is sensitive only to the values of $q_m$, $k_n$, and their relative position $(m-l)$.

In practice, new vectors $\tilde{q_m}$ and $\tilde{k_l}$ are computed following this specific equation
\begin{equation}
\begin{split}
\tilde{q_m}=R^m_{\theta,d}(q_m) \\
\tilde{k_l}=R^l_{\theta,d}(k_l)
\end{split}
\end{equation}
with 
\begin{equation}
\begin{multlined}
 R^d_{\Theta, m}(x) = 
	\begin{pmatrix}
		x_1\\
		x_2\\
		x_3\\
		x_4\\
		\vdots\\
		x_{d-1}\\
		x_d
	\end{pmatrix}
	\otimes
	\begin{pmatrix}
		\cos{m\theta_1} \\
		\cos{m\theta_1} \\
		\cos{m\theta_2} \\
		\cos{m\theta_2} \\
		\vdots \\
		\cos{m\theta_{d/2}} \\
		\cos{m\theta_{d/2}} 
	\end{pmatrix}
	+ \\
	\begin{pmatrix}
		-x_2\\
		x_1\\
		-x_4\\
		x_3\\
		\vdots\\
		-x_{d-1}\\
		x_d
	\end{pmatrix}
	\otimes
	\begin{pmatrix}
		\sin{m\theta_1}\\
		\sin{m\theta_1}\\
		\sin{m\theta_2}\\
		\sin{m\theta_2}\\
		\vdots\\
		\sin{m\theta_{d/2}}\\
		\sin{m\theta_{d/2}}
	\end{pmatrix}
 \end{multlined}
 \end{equation} 

where $m$ is the indice position and $d$ is the dimension of the $x$.

While RoPE is in theory expandable to any sequence lengths, we found an exploding perplexity when directly extending a pre-trained model beyond the context length \(L\) of the pretraining process\cite{chen2023extendingPI}. In order to overcome this problem, we interpolate position indices from longer context length \(L'\) (i.e., \([0, L')\)) to original pre-trained context length \(L\) (i.e., \([0, L)\)) in order to match the original range of indices.

Formally, we replace $R^d_{\Theta, m}(x)$  by ${R^{'d}_{\Theta, m}}(x)$ function where 
\begin{equation}
     {R^{'d}_{\Theta, m}}(x)=R^d_{\Theta, m \cdot \alpha}(x)
\end{equation}
with a parameterized scaling factor $\alpha$ defined as below: \[\alpha=\dfrac{L}{L'}\]

\subsection{Self-Attention Module and Output Projection}

Once $\{\tilde{q},\tilde{k},v\}$ are computed, we compute the outputs $o$ via a self-attention module as
\begin{equation}
    o=softmax(\tilde{q}\tilde{k}^T)v
\end{equation}
A significant challenge associated with self-attention is its computational burden when handling long sequences. Specifically, both computational and memory requirements increase quadratically with the length of the sequence. To address this limitation, we replace the original attention computation with Flash Attention 2\cite{dao2023flashattention2}, an Input-Output attention algorithm that scales memory consumption linearly and accelerates the training process. It's important to note that the attention computed using Flash Attention 2 is identical to that of the original operation. In practical terms, this allows us to process sequences of up to 12,000 tokens on a single A10 GPU card.

Subsequently, these outputs are projected by a linear layer with a weight matrix $W_o$. Similarly to the projection matrices $\{W_q , W_k , W_v\}$ We apply LoRA to the output projection matrix $W_o$.

\subsection{Mean Pooling and Classification Layer}

Unlike BERT models, LLMs such as LLAMA2 or ChatGLM2 do not use a 'CLS' token like in other models at the beginning of each sentence to get its representation. Nonetheless, we think that some knowledge is still well encoded within the token representations and a mean pooling across all input sentence tokens should provide a natural sentence representation\cite{ni-etal-2022-sentence}. Therefore, we apply a mean pooling at the sentence level. More precisely, for each sentence $s_i$ comprised of a list of $M_i$ contextualized words processed by previous decoder layers of the LLM $\{w_{i,1},w_{i,2} \dots w_{i,M_i}\}$, we compute $\bar{s_i}$
\begin{equation}
\bar{s_i} = \frac{1}{M_i} \sum_{j=1}^{M_i} w_{i,j}    
\end{equation}

Once, $\bar{s_i}$ obtained, we pass it through a linear classification layer :

\begin{equation}
    \hat{y_i} = \sigma(W_c \bar{s_i} + b)
\end{equation}

where $W_c$ is a weight matrix trainable, $b$ is a biais term trainable and $\sigma$ is the sigmoid function. The loss function used is the binary cross entropy between $\hat{y_i}$ and the oracle $y_i$.

\section{Experiments}

In this section, we present the results of our various experiments demonstrating the performances of our models in different settings compared to strong baselines of the state-of-the-art. 

\subsection{Datasets}

\begin{table*}[t]
    \small
    \centering
    \begin{tabular}{|c|c|c|c|c|c|c|} 
        \hline
        Set& \multicolumn{3}{|c|}{PubMed} & \multicolumn{3}{|c|}{Arxiv} \\ 
        \hline
        & Length-4K& Length-12K& Length-16K& Length-4K& Length-12K& Length-32K\\ 
        \hline
         Train&  70893&  127192&  131233&  38532&  153802&  202648\\ 
        \hline
         Validation&  3630&  6442&  6630&  1124&  5089&  6435\\
        \hline
         Test&  3682&  6472&  6657&  1076&  5085&  6439\\ \hline
    \end{tabular}
    \caption{The datasets we used in the experiments. cell values correspond to the number of documents for each dataset and split. Pubmed-16k and Arxiv-32k have been truncated from original datasets, contrary to the other datasets where longer documents have been filtered out.}
    \label{table:datasets}
\end{table*}

For our experiments, we evaluate our approach with two sources wildly used in summarization tasks, namely the arXiv and PubMed datasets\cite{cohan-etal-2018-discourse}. They consist of rather long scientific papers, with PubMed focusing on the biomedical domain while arXiv includes articles from various scientific fields.

To train the model in a supervised manner for the extractive summarization task, sentence labels are needed. We use the already-computed labels from \cite{cho-etal-2022-toward}\footnote{https://github.com/tencent-ailab/Lodoss/tree/main}. They followed the methodology of \cite{kedzie-etal-2018-content} with the objective of maximizing the average of the R1 and R2 scores. Moreover, we derive from the original dataset two filtered datasets containing only documents shorter that a given sequence length, in order to evaluate EYEGLAXS trained on shorter documents when tested on longer documents.
Characteristics of the resulting datasets are shown in the Table \ref{table:datasets}.


\subsection{Experimental settings}
\label{section:experimental_setting}

We used and modified the implementation released on the TransformerSum\footnote{https://github.com/HHousen/TransformerSum/tree/master}. Experiments have been carried out on 8 NVIDIA A10G GPUs. Aside from the experiments on low-volume data, models have been trained for 5 epochs, with a validation step occurring every fifth of an epoch. Models were saved based on the smallest validation loss achieved. We use a batch size of 1 with gradient accumulation every 32 steps and the adam8bit optimizer with a 3e-5 learning rate. Gradient-checkpointing and bf16-mixed precision are used. Deepspeed stage 1 is employed. No advanced hyperparameter search was performed. We use sequence lengths of 4k and 12k to train our models before testing them on the full length dataset. Results are obtained without trigram blocking by selecting the 7 and 5 sentences with the highest probability scores for the PubMed and arXiv datasets respectively as it is done in \cite{cho-etal-2022-toward}. The scaling factor $\alpha$ for RoPE is set to 8 to handle up to 32K length context. Rank $r$ of LoRA is set to 8. 

\subsection{Evaluation Metrics}

We use ROUGE scores to evaluate the model performance\cite{lin-2004-rouge}. More precisely, we report the F1 score of unigram, bigram overlap (ROUGE-1, ROUGE-2) and the longest common subsequence (ROUGE-L). We use the python implementation\footnote{https://github.com/google-research/google-research/tree/master/rouge}.

\subsection{Baseline systems}

\begin{table}
    \centering
    \small
    \begin{tabular}{|l|c|c|c|} \hline 
         Models&  R-1&  R-2& R-L\\ \hline 
         \hline
         \multicolumn{4}{|c|}{Abstractive Models}\\ \hline 
         Bigbird-large&46.32&20.65&42.33\\ \hline 
         Long-T5&50.23  & 24.76 & 46.67 \\ \hline 
         \hline
         \multicolumn{4}{|c|}{Extractive Models}\\ \hline 
         ORACLE& 61.49 & 34.70 & 55.92 \\ \hline 
 LEAD-10& 37.45&14.19 &34.07\\ \hline 
 SumBasic& 37.15&11.36 &33.43\\ \hline 
 LexRank&39.19&13.89 &34.59 \\ \hline 
 Sent-PTR& 45.01&19.91 &41.16\\ \hline 
 GenCompareSum&42.10 &16.51 &38.25\\ \hline 
 Histruct+&46.59 & 20.39&42.11\\ \hline 
 Lodoss-base (Longformer)& 48.10&22.53 &43.51\\ \hline 
 Lodoss-full-LG& 49.38&23.89 &44.84\\ \hline 
 GoSum&49.83 &23.56 &45.10\\\hline \hline 
 \hline
 \multicolumn{4}{|c|}{Our system EYEGLAXS (Extractive)}\\ \hline 
 CHATGLM2-6B (4K)&49.96 &24.04 &45.50\\ \hline 
 CHATGLM2-6B (12K)&50.17 &24.41 &45.66\\ \hline 
 LLAMA2-7B (4K)&49.48 &23.64 &45.08\\ \hline 
 LLAMA2-7B (12K)&\textbf{50.34 }&\textbf{24.57} &\textbf{45.96}\\ \hline
    \end{tabular}
    \caption{ROUGE results on the PubMed dataset}
    \label{tab:result_pubmed}
\end{table}

Using the arXiv and PubMed datasets, which are two popular datasets in the domain of extractive summarization, allows us to easily assess the relevance of our approach. We can directly compare the results obtained with our models against previous systems of the state-of-the-art on the same ROUGE metrics. 
Among the baselines, we compare our approach with standard lexical methods like Sumbasic \cite{vanderwende2007beyond} that is based on word frequencies or LexRank \cite{erkan2004lexrank} that uses a graph-based approach and centrality scoring of sentences. Taking advantage of what language models have to offer and building upon the BERTSUM framework \cite{liu-lapata-2019-text}, we have strong extractive baselines: HiStruct+ \cite{ruan-etal-2022-histruct} explicitly exploits the hierarchical structure of the text, taking advantage of the position of sentences within sections. Among more recent models, Lodoss \cite{cho-etal-2022-toward} represents a strong baseline that achieved great performances by being jointly trained for text summarization and segmentation in addition to using a novel regularizer to boost diversity among the selected summary sentences. GoSum \cite{bian2023gosum} is another state-of-the-art model that showed some of the best results by exploiting graph neural networks and reinforcement learning. Alongside these extractive models that we directly compare ourselves to, we also show the performances of some popular abstractive baseline. Rather than measuring up against them, we simply add them as a reference to put the results into perspective. 



\section{Results and Analyses}

\begin{table}
    \centering
    \small
    \begin{tabular}{|c|c|c|c|} \hline 
         Models&  R-1&  R-2& R-L\\ \hline 
         \multicolumn{4}{|c|}{Abstractive Models}\\ \hline 
         Bigbird-large&46.63&19.02&41.77\\ \hline 
         Long-T5& 48.35 & 21.92 & 44.27 \\ \hline 
         \multicolumn{4}{|c|}{Extractive Models}\\ \hline 
         ORACLE& 59.41 & 30.05 & 52.34 \\ \hline 
 LEAD-10&35.52 &10.33 &31.44\\ \hline 
 SumBasic&29.47 &6.95&26.30\\ \hline 
 LexRank&33.85 &10.73 &28.99\\ \hline 
 Sent-PTR& 42.32& 15.63&38.06\\ \hline 
 GenCompareSum& 39.66& 12.30&35.38\\ \hline 
 Histruct+& 45.22& 17.67&40.16\\ \hline 
 Lodoss-base (Longformer)& 47.64& 19.73&41.71\\ \hline 
 Lodoss-full-LG& 48.45& 20.72&42.55\\ \hline 
 GoSum& 48.61& 20.53&42.80\\\hline \hline 
 \multicolumn{4}{|c|}{Our system EYEGLAXS (Extractive)}\\ \hline 
 CHATGLM2-6B (4K)& 46.87&18.96 &41.37 \\ \hline 
 CHATGLM2-6B (12K)&\textbf{49.02}& 21.01&\textbf{43.33} \\ \hline  
 LLAMA2-7B (4K)& 48.68& 20.72&42.97\\ \hline 
 LLAMA2-7B (12K)&48.96 &\textbf{21.07} &43.30\\ \hline
    \end{tabular}
    \caption{ROUGE results on the arXiv dataset}
    \label{tab:result_arxiv}
\end{table}

Results on PubMed and arXiv are shown respectively in table \ref{tab:result_pubmed} and table \ref{tab:result_arxiv}. For both datasets, the EYEGLAXS variants, specifically ChatGLM2-6B (4K) and ChatGLM2-6B (12K), alongside the LLAMA variants, LLAMA2-7B (4K) and LLAMA2-7B (12K), showcase competitive performance compared to the state-of-the-art even if they are trained on smaller and shorter dataset. Models trained on a longer context (12K variant) exhibit superior performance compared to their counterpart trained on a shorter context. LLAMA2-7B (12K) and ChatGLM2-6B (12K) seem to have similar performances on both datasets. However, ChatGLM2-6B (4K) seems to underperform compared to LLAMA2-7B (4K) on arXiv, we hypothesize that since the arXiv dataset contains longer documents, LLAMA2 benefited more from his pretraining stage on long document compared to ChatGLM2 and could potentially need less amount of training data to converge (see section \ref{sec:small_datasets} ). Finally, we obtain new state-of-the-art results compared to other extractive methods on both datasets. We provide also in appendix the complete table results on filtered datasets (e.g. training LLAMA2-7B on the 4K dataset and testing it on the 12K dataset).

\subsection{Evaluating LoRA's Impact on Fine-Tuning EYEGLAXS}
\begin{table}
    \centering
    \small
    \begin{tabular}{|l|c|c|c|} \hline  
 Model& \multicolumn{3}{|c|}{4K PubMed Dataset}\\ \hline  
 & R1& R2& RL\\ \hline  
CHATGLM2-6B (4K) - Frozen& 42.79& 17.20& 38.68\\ \hline 
CHATGLM2-6B (4K) - LoRA& 49.96& 24.04& 45.50\\ \hline  
LLAMA2-7B (4K) - Frozen& 42.38& 17.12& 38.42\\ \hline 
LLAMA2-7B (4K) - LoRA& 49.48& 23.64& 45.08\\ \hline  

    \end{tabular}
    \caption{Comparison of ROUGE Scores between Frozen Weights and EYEGLAXS Models on 4K PubMed Dataset}
    \label{tab:LoRA_frozen}
\end{table}

To assess the contribution of LoRA and determine the relevance of the hidden representations provided by LLMs, we contrasted the performance of EYEGLAXS models trained on the 4K PubMed filtered dataset with a variant of same models with frozen weights, albeit with a trainable classifier head. The comparative outcomes are presented in Table \ref{tab:LoRA_frozen}. The notable enhancement in ROUGE scores underscores the necessity of employing Parameter-Efficient Fine-Tuning (PEFT) methodologies like LoRA to fine-tune LLMs, suggesting that the standalone hidden representations from LLMs may fall short of ensuring optimal performance.

\begin{figure}
    \centering
    \includegraphics[width=1\linewidth]{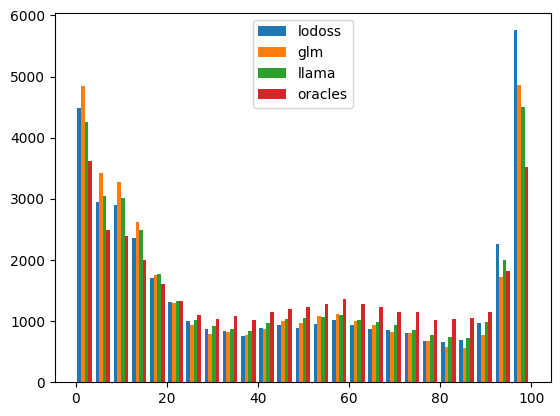}
    \caption{Number of sentences selected at each relative position by the EYEGLAXS models and baselines compared to the oracles}
    \label{fig:fig_err_1}
\end{figure}

\subsection{Position Error Analysis}

To get a better understanding of the strengths and weaknesses of our system, we further analyze the outputs of our models.
%
This section is used to check the general behavior of the model, and can help verify some reported problems with bias when using LLMs for long sequences, including a natural bias at the beginning and end of each document \cite{liu2023lost}. We choose to examine the extracted sentences and compare them to the ones forming the oracles.
To achieve this, we first trace out the distribution of the selected sentences for both versions of our model as well as a Longformer baseline (Lodoss-base) \cite{cho-etal-2022-toward} and the oracles. 
The lengths of the documents forming the datasets having a wide amplitude, we choose to use the relative positions of sentences to ensure an overall homogeneous comparison. To do so, we compare the absolute index of each extracted sentences against the document’s total number of sentences, then plot the resulting histogram. This gives us a histogram showing the relative position of the sentences selected by each model. The results are showed for the PubMed dataset on the test split. 
From the Figure \ref{fig:fig_err_1}, we can see that the sentences chosen for the oracles or predicted by the models tend to be near the beginning and the end of the document. It is not surprising since both the introduction and the conclusion usually contain sentences that are representative of the document's subject. We can observe that the three models tend to choose sentences near both ends of the text in excess. On the subject of accessing relevant information located in the middle of inputs, even though all the models are lagging behind compared to the oracles, we notice that the EYEGLAXS models follow the oracle trend a little better.

\subsection{Training on smaller datasets}
\label{sec:small_datasets}
\begin{figure}
    \centering
    \includegraphics[width=\linewidth]{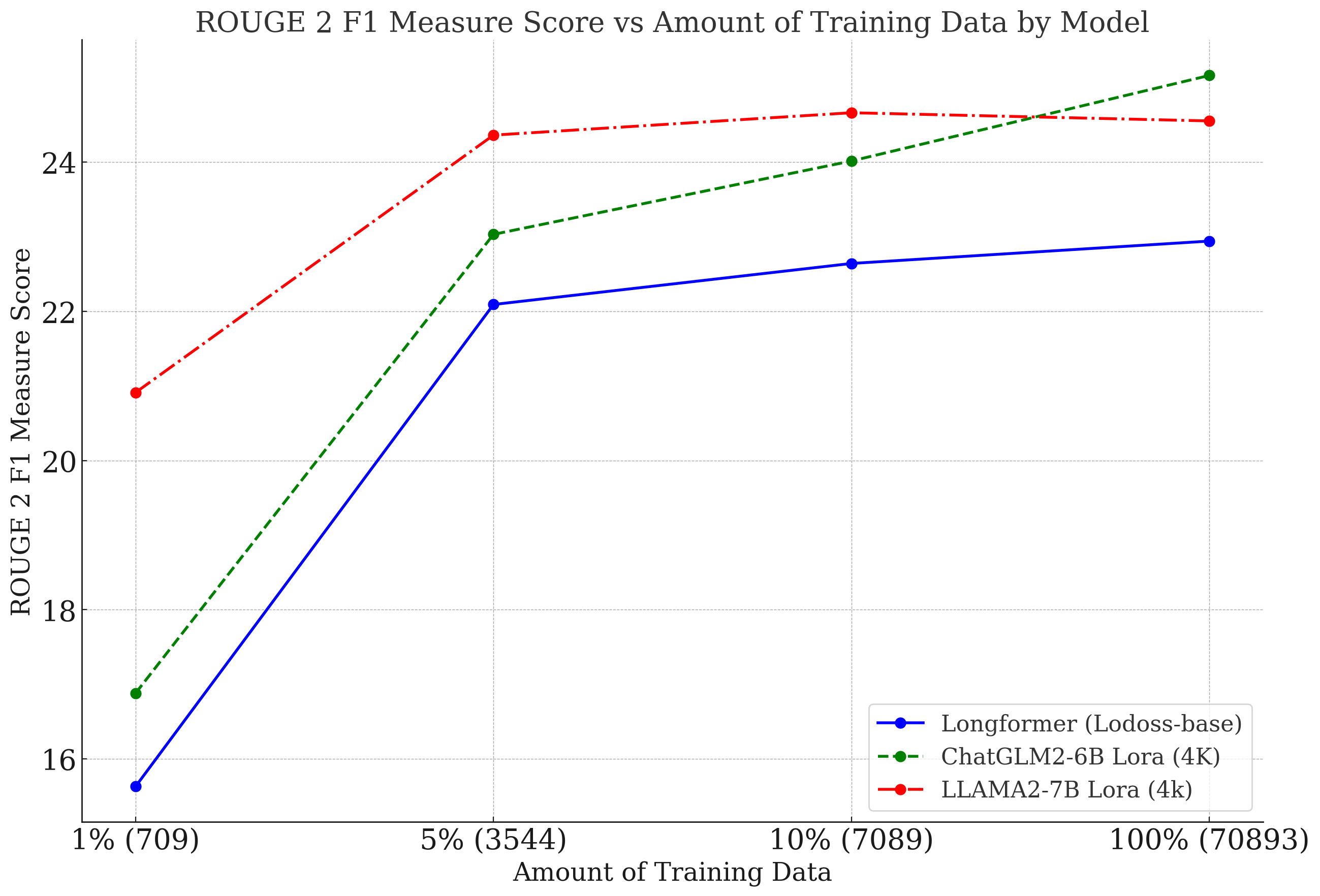}
    \caption{ROUGE-2 F1 Measure scores for Longformer (Lodoss-base), ChatGLM2-LoRA (4K) and LLAMA2-LoRA across varying training data sizes. The exact number of training instances is indicated in parentheses.}
    \label{fig:datasetsize}
\end{figure}

In the medical field, obtaining a large database like PubMed is often challenging. Assessing the performance of Large Language Models (LLMs) on smaller databases compared to traditional methods becomes crucial. To this end, we conducted an experiment comparing the performance of ChatGLM2-6B (4K), Longformer (Lodoss-base) \cite{cho-etal-2022-toward}, and LLAMA2-7B LoRA (4k), on the filtered PubMed 4K dataset using varying portions of the training data. Specifically, we examined the performance when utilizing 1\%, 5\%, 10\%, and 100\% of the PubMed 4K dataset. The findings are illustrated in Figure \ref{fig:datasetsize}.
The results confirm a positive correlation between the size of the training dataset and the performance metrics for all three models, aligning with the intuitive expectation that larger training sets generally lead to improved model performance. Interestingly, the performance gap among the three models exhibits varying dynamics as the training data size increases. 
ChatGLM2-6B (4K) and LLAMA2-7B(4K) consistently outperform Longformer across all sizes of training data, validating the efficacy of LLMs even when limited data is available. 
LLAMA2-7B LoRA (4K) starts off with a strong performance at just 1\% of the training data and maintains the lead as the dataset grows. However, its performance appears to converge, showing marginal gains as the dataset size increases compared to CHATGLM2-6B (4K) which seems to show improvements as the dataset size increases. This divergence in behavior may be attributed to the architectural differences between the two models. Specifically, ChatGLM2-6B (4K) employs bidirectional attention mechanisms, which could require a larger dataset to optimize but also offer a more favorable inductive bias for information extraction tasks.


\section{Conclusion}

This paper introduces EYEGLAXS, a novel system that leverages Large Language Models (LLMs) for long text extractive summarization. Our work challenges the traditional reliance on encoder-only models, showcases the adaptability of LLMs in managing different sequence lengths, and sets new performance standards on the PubMed and arXiv datasets. Despite these advancements, the use of LLMs comes with its own set of challenges, notably in computational resource requirements and the limitations of fine-tuning. Looking ahead, we aim to integrate sliding attention mechanisms in LLMs to further refine our system. Additionally, we plan to enrich the LLM backbone with existing techniques such as graph-based methods or Reinforcement Learning. Overall, our work paves the way for new research avenues in extractive text summarization and substantiates the utility of LLMs in this field.

\section{Limitations}
While EYEGLAXS demonstrates promising advancements in extractive text summarization, it is not without its challenges. The use of Large Language Models (LLMs) requires significant computational resources, making it less accessible for those with limited capabilities. Moreover, we report only a single run for each of our experiment due to the expensive training time, as can be seen in the Table \ref{tab:training_time} in the appendix showing the duration of the training epochs. Additionally, the sheer size of these LLMs restricts the possibility of full fine-tuning, thereby limiting further optimization and reporting upper limit of the full fine-tuning. The model's performance is also closely tied to the size of the training dataset, especially for CHATGLM2, which could be a constraint in fields where large, labeled datasets are not readily available. Lastly, the system's generalizability remains untested outside of scientific contexts like PubMed and arXiv. These limitations offer valuable avenues for future research to improve the system's robustness and applicability. Finally, we wanted to highlight again the security risk if such tool is used in sensitive applications (e.g., legal, medical), poor performance or errors could lead to serious consequences.



\bibliography{aaai24}

\appendix

\section{Results on the different datasets}

\begin{table*}
    \centering
    \small
    \begin{tabular}{|l|c|c|c|c|c|c|c|c|c|c|} \hline  
          \multicolumn{1}{|c|}{Model}&Training Context Length&  \multicolumn{9}{|c|}{Evaluation Context Length}\\ \hline  
 & & \multicolumn{9}{|c|}{ARXIV}\\ \hline  
  && \multicolumn{3}{|c|}{4K}& \multicolumn{3}{|c|}{12K}& \multicolumn{3}{|c|}{32K}\\ \hline  
 & & R1& R2& RL& R1& R2& RL& R1& R2&RL\\ \hline  
  CHATGLM2-6b&4K& 45.18& 18.07& 39.73& 47.55& 19.64& 41.97& 46.87& 18.96&41.37\\ \hline  
  &12k& 46.11& 19.10& 40.59& 48.96& 21.11& 43.26& 49.02& 21.01&43.33\\ \hline  
 LLAMA2-7b& 4K& 45.84& 18.84& 40.37& 48.58& 20.81& 42.90& 48.68& 20.72&42.98\\ \hline  
 & 12K& 45.97& 19.08& 40.54& 48.80& 21.09& 43.16& 48.96& 21.07&43.30\\ \hline  
 & & \multicolumn{9}{|c|}{PUBMED}\\ \hline  
  && \multicolumn{3}{|c|}{4K}& \multicolumn{3}{|c|}{12K}& \multicolumn{3}{|c|}{16K}\\ \hline  
 & & R1& R2& RL& R1& R2& RL& R1& R2&RL\\ \hline  
  CHATGLM2-6b&4K& 49.43& 25.17& 45.35& 50.08& 24.24& 45.64& 49.96& 24.04&45.50\\ \hline  
  &12k& 49.40& 25.21& 45.26& 50.21& 24.52& 45.72& 50.17& 24.41&45.66\\ \hline  
 LLAMA2-7b& 4K& 48.59& 24.55& 44.58& 49.54& 23.80& 45.15& 49.48& 23.64&45.08\\ \hline  
 & 12K& 50.38& 24.70& 46.02& 50.39& 24.70& 46.03& 50.34 &24.57 &45.96\\ \hline
    \end{tabular}
    \caption{ROUGE Metrics of EYEGLAXS Variants on ARXIV and PUBMED Datasets at Different Training and Evaluation Context Lengths.}
    \label{tab:context_length_complete}
\end{table*}

We provide in Table \ref{tab:context_length_complete} the different results of all variants from EYEGLAXS tested on the different versions of datasets we have built. This table provides us a better idea about how model trained on short document perfom on a dataset containing longer documents.

\section{Model Training Time}

We show in Table \ref{tab:training_time} the training time for one epoch for each model on both arXiv and PubMed. Hardware specifics and training parameters are specified in the Experimental Settings section.

\begin{table}
    \centering
    \begin{tabular}{|c|c|c|c|} \hline
        \multirow{2}{*}{Model} &Training Context&  \multicolumn{2}{|c|} {Training Time} \\ \cline{3-4}
         &Length& ARXIV & PUBMED\\ \hline
         CHATGLM2-6b & 4K & 8h 08m & 8h 06m \\ \hline
         CHATGLM2-6b & 12K & 52h 54m & 31h 14m \\ \hline
         LLAMA2-7b & 4K & 8h 36mn & 8h 33m \\ \hline
         LLAMA2-7b & 12K & 51h 35m & 32h 29m \\ \hline
    \end{tabular}
    \caption{Training Time for One Epoch of CHATGLM2-6b and LLAMA2-7b Models on ARXIV and PUBMED Datasets at Different Context Lengths.}
    \label{tab:training_time}
\end{table}

\end{document}